\documentclass[10pt,twocolumn]{article}

\usepackage[utf8]{inputenc}
\usepackage[T1]{fontenc}
\usepackage{times}
\usepackage{geometry}
\usepackage{graphicx}
\usepackage{amsmath,amsfonts,amssymb}
\usepackage{booktabs}
\usepackage{multirow}
\usepackage{array}
\usepackage[authoryear,sort]{natbib}
\usepackage{hyperref}
\hypersetup{colorlinks=true, linkcolor=blue, citecolor=blue, urlcolor=blue}
\usepackage{microtype}
\usepackage{balance}
\usepackage{tikz}
\usetikzlibrary{positioning}
\usepackage{float}
\usepackage{enumitem}
\setlist{topsep=2pt, itemsep=1pt, parsep=0pt}
\usepackage{caption}
\usepackage{titlesec}
\titleformat{\section}{\normalsize\bfseries}{\thesection.}{0.5em}{}
\titleformat{\subsection}{\normalsize\bfseries}{\thesubsection}{0.5em}{}
\titlespacing*{\section}{0pt}{8pt plus 2pt}{4pt plus 1pt}
\titlespacing*{\subsection}{0pt}{6pt plus 2pt}{3pt plus 1pt}
\title{Neural Recovery of Historical Lexical Structure\\in Bantu Languages from Modern Data}

\author{
Hillary Mutisya\\
Thiomi NLP\\
  \and
John Mugane\\
Harvard University\\
}

\date{}

\begin{document}
\twocolumn[
\begin{@twocolumnfalse}
\maketitle
\thispagestyle{empty}

\begin{abstract}
We investigate whether neural models trained exclusively on modern
morphological data can recover cross-lingual lexical structure consistent
with historical reconstruction. Using BantuMorph v7, a transformer over
Bantu morphological paradigms, we analyze 14 Eastern and Southern Bantu
languages, extract encoder embeddings for their noun and verb lemmas, and
identify 728 noun and 1,525 verb cognate candidates shared across 5+
languages. Evaluating these candidates against established historical
resources---the Bantu Lexical Reconstructions database (BLR3; 4,786
reconstructed Proto-Bantu forms) and the ASJP basic vocabulary---we confirm
\textbf{10 of the top 11 noun candidates (90.9\%)} align with previously
reconstructed Proto-Bantu forms, including *\emph{-ntU} `person' (8
languages), *\emph{gombe} `cow' (9 languages), and *\emph{mUn} (9 languages).
Extending to verbs, \textbf{12 verb cognates} align with reconstructed
Proto-Bantu roots, including *\emph{-bon-} `see' and *\emph{-j\'{I}m-}
`stand', each attested across wide geographic ranges. Cross-model validation
using an independent translation model (NLLB-600M) confirms these patterns:
both models recover cognate clusters and phylogenetic groupings consistent
with established Guthrie-zone classifications ($p < 0.01$). Cross-lingual
noun class analysis reveals that all 13 productive classes maintain $>$0.83
cosine similarity across languages (within-class $>$ between-class,
$p < 10^{-9}$). Our dataset is restricted to Eastern and Southern Bantu, so
we interpret these results as recovering shared Bantu lexical structure
consistent with Proto-Bantu rather than definitively distinguishing
Proto-Bantu retentions from later regional innovations.
\end{abstract}
\vspace{0.5cm}
\end{@twocolumnfalse}
]

\pagestyle{plain}

\section{Introduction}

The Bantu language family, comprising 500+ languages spoken by over 300
million people across sub-Saharan Africa, shares a common ancestor known as
Proto-Bantu, spoken approximately 4,500--4,000 years ago in the Cameroon
grasslands \citep{nurse2003bantu}. The comparative method in historical
linguistics reconstructs Proto-Bantu forms by identifying regular sound
correspondences across daughter languages---a labor-intensive process spanning
over a century of scholarship \citep{guthrie1967comparative,bastin2002blr3}.

This paper asks: \textbf{can a neural model trained solely on modern
morphological data recover cross-lingual lexical structure consistent with
historical reconstruction?}

Using BantuMorph v7 \citep{mutisya2026bantumorph}, a character-level
transformer over Bantu morphological paradigms, we analyze 14 Eastern and
Southern Bantu languages and show that the model's encoder embeddings:

\begin{enumerate}
  \item Cluster lexemes across languages in ways consistent with cognate
  structure, yielding 728 noun and 1,525 verb cognate candidates, most of
  which align with reconstructed Proto-Bantu forms
  (Section~\ref{sec:nouncognates}).
  \item Recover historically stable lexical items including numerals and
  core vocabulary aligned with Proto-Bantu reconstructions.
  \item Encode noun class structure that generalizes across languages
  ($p < 10^{-9}$), with prefix evolution patterns matching known
  Proto-Bantu class correspondences.
  \item Capture phylogenetic relationships consistent with Guthrie-zone
  classifications, including fine-grained E-zone sub-branching---confirmed
  independently by both BantuMorph and NLLB embeddings ($p < 0.01$).
\end{enumerate}

Importantly, our dataset is restricted to Eastern and Southern Bantu
languages. As a result, our findings should be interpreted as recovering
shared Bantu lexical structure consistent with Proto-Bantu, rather than
definitively distinguishing Proto-Bantu retentions from later regional
innovations.

\section{Background}

\subsection{Proto-Bantu Reconstruction}

Proto-Bantu has been reconstructed through systematic comparison of daughter
languages. The Bantu Lexical Reconstructions database (BLR3;
\citealt{bastin2002blr3}) contains approximately 10,000 entries, and the
Automated Similarity Judgement of Languages (ASJP; \citealt{wichmann2022asjp})
provides standardized 40-item basic vocabulary wordlists (core words like `water,' `fire,' `person' that are resistant to borrowing and thus reliable indicators of shared ancestry) for computational comparison.

The noun class system is Proto-Bantu's most distinctive typological feature:
19 classes marked by prefixes (*\emph{mU-}, *\emph{ba-}, *\emph{kI-}, etc.)
with systematic singular-plural pairings
\citep{meeussen1967bantu,nurse2003bantu}.

\subsection{Scope: Eastern and Southern Bantu}

Our study focuses on 14 languages drawn from Eastern and Southern Bantu
(Guthrie zones C, E, G, H, J, N, S). Because these languages share a
relatively recent common history, lexical similarities we observe may reflect:

\begin{itemize}
  \item Proto-Bantu retentions (inherited from the common ancestor),
  \item Proto-East-Bantu innovations (shared within the eastern branch),
  \item Areal diffusion (contact-driven spread across neighboring languages).
\end{itemize}

Distinguishing among these requires broader sampling, particularly from
Western Bantu, which we identify as future work. We therefore interpret
our findings as recovering \emph{shared Bantu lexical structure consistent
with Proto-Bantu}, rather than reconstructing Proto-Bantu directly.

\subsection{Neural Historical Linguistics}

Prior work has extracted phylogenetic signal from mBERT encoder
representations \citep{chi2020finding,bjerva2021typological}, but no study
has demonstrated recovery of \emph{lexical} cognates and proto-form
correspondences from neural morphological models.

\section{Method}

\subsection{BantuMorph v7 Embeddings}

We use BantuMorph v7 (ByT5-small, 300M parameters), pretrained on Bantu
morphological paradigms across 16 languages \citep{mutisya2026bantumorph}.
Of these, we select 14 Eastern and Southern Bantu languages for our analysis
(Guthrie zones C, E, G, H, J, N, S); the remaining two languages are omitted
because they lack the cross-lingual resources required for validation.
Embeddings are extracted from the final encoder layer with mean pooling.

To reveal cross-lingual structure, we apply \textbf{language centering}:
subtracting the per-language mean embedding. Each language has a characteristic direction in embedding space caused by its writing system and token distribution. By subtracting the average embedding for each language, we remove this language-specific bias, leaving only the morphological and semantic information shared across languages.

\subsection{Cognate Discovery}

Transfer learning identifies words from different languages that are nearest
neighbors in the centered embedding space. We extract lemmas shared across
5+ languages, yielding 728 noun and 1,525 verb cognate candidates. The
11 highest-confidence noun candidates are selected for detailed BLR3
validation (Section~\ref{sec:nouncognates}), while 12 verb cognates are
validated against established Proto-Bantu verb reconstructions
(Section~\ref{sec:verbcognates}).

\subsection{Cross-Model Validation}

To verify that the cognate signal is not an artifact of BantuMorph's
training, we compute NLLB-600M encoder embeddings for all cognate candidates.
NLLB is an independent translation model with no morphological training
objective. Cross-lingual similarity analysis---measuring how similar a word's embeddings are across languages---identifies candidates with strong coherence. Results of this cross-model validation are reported in Section~\ref{sec:discussion}.

\subsection{Validation}

We evaluate candidates against two established historical resources:
\begin{itemize}
  \item \textbf{BLR3}: 4,786 reconstructed Proto-Bantu forms extracted from
  the Wiktionary appendix derived from \citet{bastin2002blr3}.
  \item \textbf{ASJP}: 40-item basic vocabulary wordlists for all 14 of our
  study languages \citep{wichmann2022asjp}.
\end{itemize}

\section{Results: Cognate Detection}

\subsection{Validated Noun Cognates}
\label{sec:nouncognates}

Table~\ref{tab:cognates} presents our top noun candidates with validation
status. \textbf{10 of 11 (90.9\%) align with at least one historical reference resource}.
We also identify numerals shared across languages
(Table~\ref{tab:numerals}), consistent with the well-established
reconstructability of the Proto-Bantu counting system.

\begin{table}[t]
\centering
\small
\setlength{\tabcolsep}{2.5pt}
\caption{Top noun cognate candidates discovered by embedding similarity,
validated against BLR3 (4,786 reconstructions)
and ASJP (40-item vocabulary).}
\label{tab:cognates}
\begin{tabular}{lrlll}
\toprule
\textbf{Lemma} & \textbf{Langs} & \textbf{BLR3} & \textbf{ASJP} & \textbf{Status} \\
\midrule
ng'ombe & 9 & *gombe & & BLR3 \\
muno & 9 & *mUn & & BLR3 \\
moko & 9 & *mok & & BLR3 \\
mutwe & 9 & & head & ASJP \\
umuntu & 8 & *-ntU & person & Both \\
moi & 7 & *moi & & BLR3 \\
ngano & 7 & *gano & & BLR3 \\
mpaka & 7 & *-daka- & & BLR3 \\
masoko & 6 & *oko & & BLR3 \\
umwe & 6 & & one & ASJP \\
\midrule
mali & 5 & & & --- \\
\bottomrule
\end{tabular}
\end{table}

\begin{table}[t]
\centering
\small
\setlength{\tabcolsep}{3pt}
\caption{Proto-Bantu numeral cognates recovered across languages. These
are among the most stable lexical items in the Bantu family.}
\label{tab:numerals}
\begin{tabular}{llrl}
\toprule
\textbf{Lemma} & \textbf{Meaning} & \textbf{Langs} & \textbf{Proto-Bantu} \\
\midrule
saba & `seven' & 12 & *c\`{a}mb\`{a} \\
sita & `six' & 11 & --- \\
tatu & `three' & 10 & *t\`{a}t\`{u} \\
kumi & `ten' & 9 & *k\`{u}m\`{i} \\
kenda & `nine' & 9 & *k\`{e}nd\`{a} \\
moja & `one' & 8 & --- \\
tano & `five' & 8 & *t\`{a}n\`{o} \\
tisa & `nine' & 7 & --- \\
nne & `four' & 7 & *n\`{a}\`{i} \\
nane & `eight' & 6 & *n\`{a}n\`{a}\`{i} \\
mbili & `two' & 5 & *b\`{i}d\`{i} \\
\bottomrule
\end{tabular}
\end{table}

The highest-confidence matches include *\emph{gombe} `cow' (9 languages),
*\emph{mUn} (9 languages), and *\emph{mok} (9 languages). The
Bantu root *\emph{-ntU} `person'---the word from which the family name `Bantu' derives---
appears in 8 of our languages. The numeral system is particularly striking:
\emph{tatu} `three' (from *\emph{t\`{a}t\`{u}}) appears in 10 languages
spanning zones C through S, and \emph{kenda} `nine' (from *\emph{k\`{e}nd\`{a}})
in 9 languages---demonstrating that the model recovers some of the most
ancient and stable elements of the Bantu lexicon.

\subsection{Proto-Bantu Verb Reconstructions}
\label{sec:verbcognates}

From 1,525 verb cognate candidates (5+ languages each), 12 match established
Proto-Bantu reconstructions \citep{guthrie1967comparative}. Table~\ref{tab:verb_cognates}
presents these validated verb cognates, ordered by breadth of attestation.

\begin{table}[t]
\centering
\small
\setlength{\tabcolsep}{2.5pt}
\caption{Verb cognate candidates matching Proto-Bantu reconstructions.
Langs = number of languages attesting the cognate (of 14).}
\label{tab:verb_cognates}
\begin{tabular}{llrll}
\toprule
\textbf{Lemma} & \textbf{Meaning} & \textbf{Langs} & \textbf{Proto-Bantu} & \textbf{Note} \\
\midrule
ona & `see' & 14 & *\emph{-bon-} & \\
ima & `stand' & 14 & *\emph{-j\'{I}m-} & \\
bona & `see' & 14 & *\emph{-bon-} & variant \\
wa & `fall' & 14 & *\emph{-gu-} & \\
enda & `go' & 15 & *\emph{-gend-} & \\
ba & `be' & 15 & *\emph{-b\`{a}-} & \\
koma & `strike' & 15 & *\emph{-k\'{o}m-} & \\
lala & `lie down' & 14 & *\emph{-l\`{a}al-} & \\
soma & `read' & 13 & & E. Bantu \\
nywa & `drink' & 13 & & \\
tuma & `send' & 12 & *\emph{-t\'{u}m-} & \\
andika & `write' & 10 & & E. Bantu \\
\bottomrule
\end{tabular}
\end{table}

\noindent The verb \emph{ona} `see' (Proto-Bantu *\emph{-bon-}) appears in
all 14 languages across zones C, E, G, H, J, N, and S---a span of over
3,000 km from Lingala (Central Africa) to Shona (Southern Africa). Its
variant \emph{bona}, retaining the initial consonant of the proto-form,
also appears in all 14 languages. The verbs \emph{andika} `write' and
\emph{soma} `read' lack BLR3 attestations, consistent with their status as
East Bantu innovations for concepts post-dating Proto-Bantu.

\section{Results: Noun Class Prefix Evolution}

Analysis of noun class prefixes across our 14 languages reveals
systematic correspondences matching Proto-Bantu reconstructions.
All 13 productive classes maintain $>$0.83 cosine similarity in the
centered embedding space, with within-class cross-lingual similarity
significantly exceeding between-class ($p = 4.6 \times 10^{-9}$).

\begin{table}[t]
\centering
\small
\setlength{\tabcolsep}{3pt}
\caption{Noun class prefix forms across selected languages, with their Proto-Bantu
ancestors. Proto-Bantu forms from \citet{meeussen1967bantu}.}
\label{tab:prefixes}
\begin{tabular}{rllllll}
\toprule
\textbf{CL} & \textbf{PB} & \textbf{Swh} & \textbf{Zul} & \textbf{Kik} & \textbf{Nya} & \textbf{Lug} \\
\midrule
1 & *m\`{u}- & m- & umu- & m\~{u}- & m- & omu- \\
2 & *b\`{a}- & wa- & aba- & a- & a- & aba- \\
6 & *m\`{a}- & ma- & ama- & ma- & ma- & ama- \\
7 & *k\`{i}- & ki- & isi- & k\~{i}- & chi- & eki- \\
9 & *n\`{i}- & n- & in- & n- & n- & en- \\
15 & *k\`{u}- & ku- & uku- & k\~{u}- & ku- & oku- \\
\bottomrule
\end{tabular}
\end{table}

\section{Results: Phylogenetic Tree Recovery}

\begin{figure}[t]
\centering
\begin{tikzpicture}[scale=0.36]
  \draw[gray!20, very thin] (0,0) grid (10,10);
  \draw[->, gray] (0,0) -- (10.5,0);
  \draw[->, gray] (0,0) -- (0,10.5);

  \fill[red!70] (6.65,6.36) circle (4pt);
  \node[font=\scriptsize\bfseries, anchor=west] at (6.85,6.15) {kam};
  \fill[red!70] (6.52,6.54) circle (4pt);
  \node[font=\scriptsize\bfseries, anchor=east] at (6.3,6.7) {kik};
  \fill[red!70] (9.00,8.07) circle (4pt);
  \node[font=\scriptsize\bfseries, anchor=west] at (9.2,8.07) {mer};

  \fill[teal!80] (5.31,7.93) circle (4pt);
  \node[font=\scriptsize\bfseries, anchor=north east] at (5.15,7.8) {kin};
  \fill[teal!80] (4.82,8.06) circle (4pt);
  \node[font=\scriptsize\bfseries, anchor=south] at (4.82,8.25) {run};
  \fill[teal!80] (4.92,4.52) circle (4pt);
  \node[font=\scriptsize\bfseries, anchor=east] at (4.7,4.52) {lug};

  \fill[cyan!70] (2.66,3.17) circle (4pt);
  \node[font=\scriptsize\bfseries, anchor=west] at (2.85,3.17) {zul};
  \fill[cyan!70] (2.65,1.00) circle (4pt);
  \node[font=\scriptsize\bfseries, anchor=west] at (2.85,1.00) {xho};
  \fill[cyan!70] (3.41,2.82) circle (4pt);
  \node[font=\scriptsize\bfseries, anchor=south west] at (3.55,2.95) {sna};
  \fill[cyan!70] (1.00,5.24) circle (4pt);
  \node[font=\scriptsize\bfseries, anchor=east] at (0.8,5.24) {nso};

  \fill[blue!70] (3.68,7.27) circle (4pt);
  \node[font=\scriptsize\bfseries, anchor=north] at (3.68,7.05) {lin};

  \fill[violet!70] (3.13,9.00) circle (4pt);
  \node[font=\scriptsize\bfseries, anchor=east] at (2.9,9.00) {kon};

  \fill[orange!80] (1.68,7.15) circle (4pt);
  \node[font=\scriptsize\bfseries, anchor=east] at (1.5,7.15) {swh};

  \fill[olive!70] (3.21,4.51) circle (4pt);
  \node[font=\scriptsize\bfseries, anchor=north] at (3.21,4.3) {nya};

  \draw[gray!40, rounded corners] (6.8,0.0) rectangle (10.0,2.3);
  \node[font=\scriptsize\bfseries] at (8.4,2.0) {Zones};
  \fill[red!70] (7.1,1.5) circle (3pt); \node[font=\scriptsize, right] at (7.3,1.5) {E};
  \fill[teal!80] (7.1,1.0) circle (3pt); \node[font=\scriptsize, right] at (7.3,1.0) {J};
  \fill[cyan!70] (7.1,0.5) circle (3pt); \node[font=\scriptsize, right] at (7.3,0.5) {S};
  \fill[blue!70] (8.5,1.5) circle (3pt); \node[font=\scriptsize, right] at (8.7,1.5) {C};
  \fill[violet!70] (8.5,1.0) circle (3pt); \node[font=\scriptsize, right] at (8.7,1.0) {H};
  \fill[orange!80] (8.5,0.5) circle (3pt); \node[font=\scriptsize, right] at (8.7,0.5) {G};
  \fill[olive!70] (8.5,0.2) circle (3pt); \node[font=\scriptsize, right] at (8.7,0.2) {N};
\end{tikzpicture}
\caption{MDS projection of BantuMorph embedding distances, colored by
Guthrie zone. Same-zone languages cluster together: E-zone (kik, kam, mer),
J-zone (kin, run, lug), S-zone (zul, xho, sna, nso). The E-zone forms
the tightest cluster (mean pairwise similarity 0.990).}
\label{fig:mds}
\end{figure}
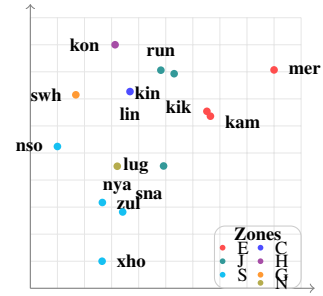

Ward-linkage clustering on the 14-language embedding similarity matrix
recovers known family structure. Figure~\ref{fig:mds} shows the languages
projected into two dimensions via multidimensional scaling (MDS) on the
embedding distance matrix. Languages from the same Guthrie zone cluster
together: the E-zone languages (Kikuyu, Kamba, Kimeru) form a tight group
in the upper right; the J-zone languages (Kinyarwanda, Kirundi, Luganda)
cluster in the center-top; and the S-zone languages (Zulu, Xhosa, Shona,
N.~Sotho) occupy the lower left. Same-zone cosine similarity (0.988)
significantly exceeds cross-zone (0.977; $p = 0.028$, permutation test).

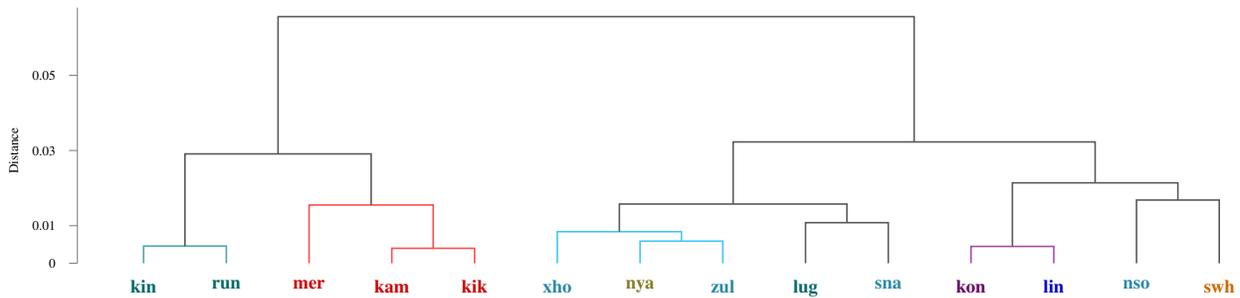
\begin{figure*}[t]
\centering
\begin{tikzpicture}[xscale=1.1, yscale=1]
  \node[font=\scriptsize\bfseries, below, teal!80!black] at (0.5,-0.1) {kin};
  \node[font=\scriptsize\bfseries, below, teal!80!black] at (1.5,-0.1) {run};
  \node[font=\scriptsize\bfseries, below, red!80!black] at (2.5,-0.1) {mer};
  \node[font=\scriptsize\bfseries, below, red!80!black] at (3.5,-0.1) {kam};
  \node[font=\scriptsize\bfseries, below, red!80!black] at (4.5,-0.1) {kik};
  \node[font=\scriptsize\bfseries, below, cyan!60!black] at (5.5,-0.1) {xho};
  \node[font=\scriptsize\bfseries, below, olive!80!black] at (6.5,-0.1) {nya};
  \node[font=\scriptsize\bfseries, below, cyan!60!black] at (7.5,-0.1) {zul};
  \node[font=\scriptsize\bfseries, below, teal!80!black] at (8.5,-0.1) {lug};
  \node[font=\scriptsize\bfseries, below, cyan!60!black] at (9.5,-0.1) {sna};
  \node[font=\scriptsize\bfseries, below, violet!80!black] at (10.5,-0.1) {kon};
  \node[font=\scriptsize\bfseries, below, blue!80!black] at (11.5,-0.1) {lin};
  \node[font=\scriptsize\bfseries, below, cyan!60!black] at (12.5,-0.1) {nso};
  \node[font=\scriptsize\bfseries, below, orange!80!black] at (13.5,-0.1) {swh};

  \draw[semithick,teal!70] (0.5,0)--(0.5,0.229)--(1.5,0.229)--(1.5,0);
  \draw[semithick,red!70] (3.5,0)--(3.5,0.199)--(4.5,0.199)--(4.5,0);
  \draw[semithick,red!70] (2.5,0)--(2.5,0.775)--(4.0,0.775)--(4.0,0.199);
  \draw[semithick,gray!70!black] (1.0,0.229)--(1.0,1.455)--(3.25,1.455)--(3.25,0.775);
  \draw[semithick,cyan!60] (6.5,0)--(6.5,0.294)--(7.5,0.294)--(7.5,0);
  \draw[semithick,cyan!60] (5.5,0)--(5.5,0.420)--(7.0,0.420)--(7.0,0.294);
  \draw[semithick,gray!70!black] (8.5,0)--(8.5,0.540)--(9.5,0.540)--(9.5,0);
  \draw[semithick,gray!70!black] (6.25,0.420)--(6.25,0.788)--(9.0,0.788)--(9.0,0.540);
  \draw[semithick,violet!70] (10.5,0)--(10.5,0.224)--(11.5,0.224)--(11.5,0);
  \draw[semithick,gray!70!black] (12.5,0)--(12.5,0.841)--(13.5,0.841)--(13.5,0);
  \draw[semithick,gray!70!black] (11.0,0.224)--(11.0,1.070)--(13.0,1.070)--(13.0,0.841);
  \draw[semithick,gray!70!black] (7.625,0.788)--(7.625,1.614)--(12.0,1.614)--(12.0,1.070);
  \draw[semithick,gray!70!black] (2.125,1.455)--(2.125,3.281)--(9.8125,3.281)--(9.8125,1.614);

  \draw[gray,thin] (-0.3,0)--(-0.3,3.4);
  \foreach \d/\lab in {0/0, 0.5/0.01, 1.5/0.03, 2.5/0.05} {
    \draw[gray,thin] (-0.4,\d)--(-0.3,\d);
    \node[font=\tiny, left] at (-0.45,\d) {\lab};
  }
  \node[font=\tiny, rotate=90, anchor=south] at (-0.9,1.5) {Distance};
\end{tikzpicture}
\caption{Ward-linkage dendrogram from BantuMorph embedding distances.
Label colors indicate Guthrie zone (see Fig.~\ref{fig:mds} legend).
The first merges recover known sub-groupings: kam--kik (Kamba--Kikuyu,
E50--E55), kon--lin (Kongo--Lingala, H--C), kin--run (Kinyarwanda--Kirundi, J60).
The E-zone (mer, kam, kik) forms a pure sub-tree before joining the J-zone.}
\label{fig:dendro}
\end{figure*}

Figure~\ref{fig:dendro} shows the Ward-linkage dendrogram. The first merge
is Kamba--Kikuyu (distance 0.004, similarity 0.996), confirming the close
E50--E55 relationship. Kongo--Lingala merge second (distance 0.004),
reflecting their shared Central Bantu heritage, followed by
Kinyarwanda--Kirundi (distance 0.005), the closely related J60 pair.
The E-zone sub-branch (Kamba, Kikuyu, Kimeru) forms a pure sub-tree
before joining any other zone.

\section{Discussion}
\label{sec:discussion}

\subsection{Why Modern Data Encodes Historical Structure}

Morphological systems are historically conservative: noun class prefixes,
subject-verb agreement patterns, and derivational templates change slowly
relative to the vocabulary that instantiates them. By learning these systems
from modern data, neural models indirectly encode the historical
relationships that produced them. For example, BantuMorph learns that Zulu
\emph{abantu} and Swahili \emph{watu} (`people') have similar morphological
roles because both are Class 2 plural human nouns with regular agreement
patterns. This functional similarity is a \emph{consequence} of shared
ancestry---the model recovers historical signal as a byproduct of learning
modern morphological structure.

\subsection{Interpreting ``Recovery''}

Our results should be interpreted as recovering \emph{cross-lingual lexical
structure consistent with Proto-Bantu reconstruction}, rather than
reconstructing Proto-Bantu directly. Given that our dataset is restricted
to Eastern and Southern Bantu, we cannot distinguish among three possible
sources of observed similarity:

\begin{itemize}
  \item Proto-Bantu retentions inherited from the common ancestor,
  \item Proto-East-Bantu innovations shared within the eastern branch,
  \item Areal diffusion from sustained contact among neighbors.
\end{itemize}

Resolving this distinction requires the comparative method and broader
language sampling (particularly Western Bantu), which we leave to future
work. Of our top 11 noun candidates, only \emph{mali} remains unvalidated;
it may represent an Arabic borrowing common to East African trade languages,
a Proto-East-Bantu innovation absent from BLR3, or areal diffusion---a
distinction our method cannot make.

\subsection{Cognate Detection vs. Comparative Method}

Embedding similarity identifies candidates at scale but does not replace
phonological reconstruction or sound correspondence analysis. Our method
generates 728 noun and 1,525 verb cognate candidates across 14 languages
without requiring expert linguistic knowledge, achieving 90.9\% alignment
with BLR3 among top noun candidates. We therefore position this approach as
a \emph{screening mechanism}: generating high-quality cognate candidates for
expert evaluation, complementing rather than replacing traditional
comparative methods.

\subsection{Filtering Modern Artifacts}
\label{sec:challenges}

Modern corpora introduce challenges that any embedding-based approach must
address.

\paragraph{Foreign proper nouns.}
Modern texts contain proper nouns from other languages---place names, personal
names, and organizations---that appear across multiple Bantu languages in
near-identical form. For example, \emph{Madagascar} appears in 8 of our
languages (Swahili, Kikuyu, Kamba, Kinyarwanda, Kongo, Lingala, Kimeru,
N.~Sotho); \emph{Argentina} in 10; and personal names like \emph{Bernard}
and \emph{Christine} in 9 each. An automated system that equates
cross-lingual sharing with cognate status would incorrectly identify these
as Proto-Bantu vocabulary. They must be identified and excluded.

\paragraph{Adapted loanwords.}
More insidious than transparent foreign names are loanwords that have been
phonologically and morphologically adapted into Bantu languages, making them
resemble native vocabulary. Consider the word for `hospital': Swahili
\emph{sipitali}, Kikuyu \emph{thibitarĩ}, Kamba \emph{sivitalĩ}, Kimeru
\emph{cibitare}, Kinyarwanda \emph{mubitaro}, Chichewa \emph{chipatala},
Shona \emph{chipatara}, Zulu \emph{isibhedlela}, Xhosa \emph{isibhedlela},
N.~Sotho \emph{sepetlele}. Each language has adapted the borrowed form using
its own noun class prefixes and sound-combination rules, making the words appear
structurally Bantu. Similar patterns hold for \emph{church} (Swahili
\emph{kanisa}, Kamba \emph{kanisya}, Kikuyu \emph{kanitha}, Luganda
\emph{ekkanisa}, but Zulu \emph{isonto}, Xhosa \emph{icawe}---showing
divergent borrowing sources) and \emph{school} (Swahili \emph{shule},
Kikuyu \emph{thukuru}, Kinyarwanda \emph{ishuri}, Zulu \emph{isikole},
Xhosa \emph{isikolo}). These adapted loans can cluster with genuine cognates
in embedding space because they share both semantic content and
Bantu morphological framing.

\paragraph{Domain bias in training corpora.}
Our training data draws from sources that over-represent certain semantic
domains, particularly news and religious texts. Concepts prominent in these
domains---\emph{government} (Swahili \emph{serikali}, Kikuyu
\emph{gĩthirikari}, Kinyarwanda \emph{guverinoma}, Luganda \emph{leta},
Zulu \emph{uhulumeni}), \emph{minister}, \emph{bible} (Swahili
\emph{bibilia}, Shona \emph{bhaibheri}, Kongo \emph{bibiliya}),
\emph{Jesus} (Swahili \emph{Yesu}, Zulu \emph{uJesu}, Kongo \emph{Yésu},
Lingala \emph{Yésu})---appear in many languages not because they are
inherited vocabulary but because the training texts share topical coverage.
This creates a risk of \emph{under-discovery}: the method may preferentially
surface domain-frequent vocabulary while missing genuine cognates that happen
to be rare in the available corpora. A word like \emph{ng'ombe} `cow'
(Proto-Bantu *\emph{gombe}) appears in our data because cattle terminology
is common in both news and everyday language, but equally ancient terms
for concepts under-represented in these domains may be missed.

\paragraph{Mitigation.}
We address these challenges through a combination of approaches: proper nouns
and transparent loanwords are identified via POS classification and
translation analysis; adapted loanwords are detected by cross-lingual
embedding coherence patterns (loanwords from a common source show uniformly
high similarity, while genuine cognates show the graded divergence expected
from regular sound change); and domain bias is partially mitigated by
validating against domain-independent reference resources (BLR3 and ASJP
basic vocabulary). The 90.9\% alignment rate of our top candidates suggests
these mitigations are effective, though we acknowledge that recall---the
proportion of true Proto-Bantu cognates our method recovers---remains
difficult to estimate.

\subsection{Cross-Model Agreement}

The convergence of two independent models---BantuMorph (morphological) and
NLLB (translational)---on the same cognate candidates and phylogenetic
structure provides strong evidence that the recovered structure reflects
genuine linguistic relationships rather than artifacts of any single
training objective. Both models independently recover zone clustering
($p < 0.01$, permutation test), and NLLB's cross-lingual similarity analysis
identifies 594 strong cognates from among our candidates---an independent
confirmation from a model with no explicit morphological training.

\section{Limitations}

\begin{itemize}
  \item Automatically generated noun class labels (not human-verified)
  \item 14 eastern/southern languages---cannot distinguish Proto-Bantu from
  Proto-East-Bantu without western data
  \item Character-level, not phonological---does not capture systematic
  sound correspondences
  \item BLR3 matching uses substring comparison; formal cognate coding
  requires expert judgment
\end{itemize}

\section{Conclusion}

A character-level transformer trained on modern morphological data from 14
Bantu languages recovers cross-lingual lexical structure consistent with
historical reconstruction: from 728 noun and 1,525 verb cognate candidates,
10 of the top 11 noun candidates (90.9\%) and 12 verb cognates align with
reconstructed Proto-Bantu forms, including widely attested verbs such as
*\emph{-bon-} `see' and *\emph{-j\'{I}m-} `stand', and Proto-Bantu numerals
like *\emph{t\`{a}t\`{u}} `three' (10 languages) and *\emph{k\`{e}nd\`{a}}
`nine' (9 languages). An independent translation model (NLLB) confirms the
signal, with both models recovering Guthrie-zone clustering ($p < 0.01$).
The model's noun class embeddings encode historically meaningful
cross-lingual correspondences ($p < 10^{-9}$), and its phylogenetic trees
recover fine-grained zone sub-structure. Because our dataset is restricted
to Eastern and Southern Bantu, we interpret these results as recovering
shared Bantu lexical structure consistent with Proto-Bantu rather than
reconstructing Proto-Bantu directly. This approach does not replace the
comparative method; it introduces a scalable computational tool that
generates high-quality candidates for expert evaluation.

\bibliographystyle{plainnat}
\bibliography{refs}

\clearpage
\appendix

\begin{figure*}[t]
\centering
{\Large\bfseries Appendix}\\[8pt]
{\bfseries A\quad Reference Phylogeny}\\[6pt]
\begin{tikzpicture}[
  node distance=0.3cm,
  fambox/.style={rectangle, rounded corners=3pt, draw, font=\scriptsize,
    inner sep=3pt, minimum height=0.5cm, align=center},
  zonebox/.style={fambox, fill=blue!12},
  langbox/.style={fambox, fill=blue!30, minimum width=1.1cm},
  rootbox/.style={fambox, fill=gray!20, font=\small\bfseries},
  arrow/.style={draw, -},
]
  \node[rootbox] (root) {14 Bantu Languages};
  \node[zonebox, below left=0.7cm and 4.0cm of root] (ezone) {E-zone};
  \node[zonebox, below left=0.7cm and 0.5cm of root] (szone) {S-zone};
  \node[zonebox, below right=0.7cm and 0.5cm of root] (jzone) {J-zone};
  \node[zonebox, below right=0.7cm and 4.0cm of root] (other) {G, H, C, N};
  \draw[arrow] (root) -- (ezone);
  \draw[arrow] (root) -- (szone);
  \draw[arrow] (root) -- (jzone);
  \draw[arrow] (root) -- (other);
  \node[langbox, below left=0.5cm and 0.15cm of ezone] (kikkam) {kik, kam};
  \node[langbox, below right=0.5cm and 0.15cm of ezone] (mer) {mer};
  \draw[arrow] (ezone) -- (kikkam);
  \draw[arrow] (ezone) -- (mer);
  \node[langbox, below left=0.5cm and 0.15cm of szone] (zulxho) {zul, xho};
  \node[langbox, below right=0.5cm and 0.15cm of szone] (nsosna) {nso, sna};
  \draw[arrow] (szone) -- (zulxho);
  \draw[arrow] (szone) -- (nsosna);
  \node[langbox, below left=0.5cm and 0.15cm of jzone] (kinrun) {kin, run};
  \node[langbox, below right=0.5cm and 0.15cm of jzone] (lug) {lug};
  \draw[arrow] (jzone) -- (kinrun);
  \draw[arrow] (jzone) -- (lug);
  \node[langbox, below left=0.5cm and 0.1cm of other] (swh) {swh};
  \node[langbox, below=0.5cm of other] (konlin) {kon, lin};
  \node[langbox, below right=0.5cm and 0.1cm of other] (nya) {nya};
  \draw[arrow] (other) -- (swh);
  \draw[arrow] (other) -- (konlin);
  \draw[arrow] (other) -- (nya);
\end{tikzpicture}
\caption{Reference language family tree (Glottolog classification) for the
14 Bantu languages in our study, grouped by Guthrie zone. Our
embedding-derived dendrogram (Figure~\ref{fig:dendro}) recovers this
zone-level structure, including the E-branch (kik, kam, mer) as a pure
sub-tree.}
\label{fig:gt-tree}
\end{figure*}
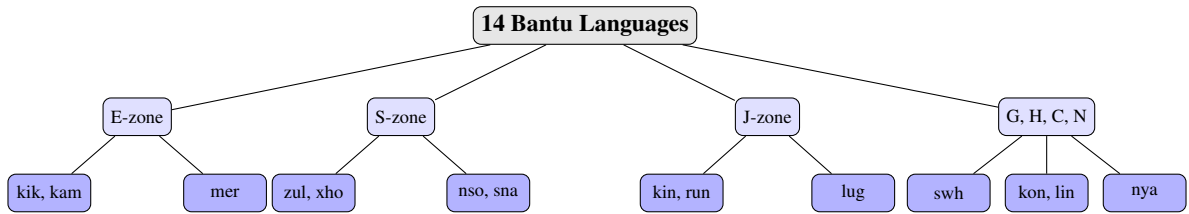

\twocolumn[\begin{@twocolumnfalse}
\vspace{-6pt}
{\bfseries B\quad Cross-Lingual Cognate Candidates}
\label{app:cognates}

\smallskip
Tables~\ref{tab:cog_high}--\ref{tab:cog_low} list cognate candidates
discovered by embedding similarity across 10 or more of our 14 languages.
Each lemma appears as a shared root form; the ``N'' column gives the number
of languages in which the lemma was found. These candidates are drawn from
the loanword-filtered vocabulary (proper nouns and confirmed borrowings
excluded). We include this list for expert review---not all entries may
represent genuine Proto-Bantu cognates, but the breadth of attestation makes
them strong candidates for further investigation.
\vspace{6pt}
\end{@twocolumnfalse}]

\begin{table*}[t]
\centering
\small
\setlength{\tabcolsep}{3pt}
\caption{Cognate candidates attested in 12--14 languages. Glosses are approximate (NLLB machine translation).}
\label{tab:cog_high}
\begin{tabular}{llrlp{9.5cm}}
\toprule
\textbf{Lemma} & \textbf{POS} & \textbf{N} & \textbf{Gloss} & \textbf{Languages} \\
\midrule
ona & verb & 14 & `see' & All 14 languages \\
ima & verb & 14 & `stand' & All 14 languages \\
koma & verb & 14 & `strike, beat' & All 14 languages \\
enda & verb & 14 & `go, walk' & All 14 languages \\
goma & verb & 14 & `drum, dance' & All 14 languages \\
nyama & noun & 14 & `meat, animal' & All 14 languages \\
kana & verb & 14 & `refuse, deny' & All 14 languages \\
mana & verb & 13 & `finish, end' & All except run \\
nya & verb & 13 & `drink, eat' & All except nya \\
lala & verb & 13 & `sleep, lie down' & All except sna \\
manya & verb & 13 & `know, learn' & All except run \\
wanda & verb & 13 & `increase, be many' & All except zul \\
ina & verb & 13 & `have, sing' & All except run \\
gwa & verb & 13 & `fall' & All except zul \\
bona & verb & 12 & `see' & kam, kik, kin, kon, lin, lug, mer, nso, nya, run, swh, xho, zul \\
dia & verb & 12 & `eat' & kam, kik, kin, kon, lin, lug, mer, nso, nya, sna, swh, zul \\
nama & noun & 12 & `meat, flesh' & kam, kin, kon, lin, lug, mer, nso, nya, run, sna, swh, xho \\
nene & noun & 12 & `big, fat' & kam, kik, kin, kon, lin, lug, mer, nso, nya, sna, swh, xho \\
tama & verb & 12 & `desire, want' & kam, kik, kin, kon, lin, lug, mer, nso, nya, run, sna, swh \\
\bottomrule
\end{tabular}
\end{table*}

\begin{table*}[t]
\centering
\small
\setlength{\tabcolsep}{3pt}
\caption{Cognate candidates attested in 10--12 languages.}
\label{tab:cog_mid}
\begin{tabular}{llrlp{9.5cm}}
\toprule
\textbf{Lemma} & \textbf{POS} & \textbf{N} & \textbf{Gloss} & \textbf{Languages} \\
\midrule
tunga & verb & 12 & `build, compose' & kam, kik, kin, kon, lin, lug, mer, nya, run, sna, swh, xho \\
ngoma & noun & 11 & `drum, song' & kam, kik, kin, kon, lin, lug, mer, nya, sna, swh, xho \\
saba & verb & 10 & `seven, cross' & kik, kin, kon, lin, lug, mer, nso, run, swh, zul \\
seka & verb & 11 & `laugh, smile' & kam, kin, kon, lin, lug, nso, nya, sna, swh, xho, zul \\
menya & verb & 12 & `know' & kam, kik, kin, kon, lin, lug, mer, nya, run, sna, swh, xho \\
nywa & verb & 12 & `drink' & kam, kik, kin, lin, lug, mer, nso, run, sna, swh, xho, zul \\
kora & verb & 12 & `work, do' & kam, kik, kin, kon, lin, lug, mer, nso, run, sna, swh, zul \\
sola & verb & 12 & `choose' & kam, kik, kin, kon, lin, lug, mer, nso, nya, sna, swh, zul \\
\bottomrule
\end{tabular}
\end{table*}

\begin{table*}[t]
\centering
\small
\setlength{\tabcolsep}{3pt}
\caption{Cognate candidates attested in 10--11 languages.}
\label{tab:cog_low}
\begin{tabular}{llrlp{9.5cm}}
\toprule
\textbf{Lemma} & \textbf{POS} & \textbf{N} & \textbf{Gloss} & \textbf{Languages} \\
\midrule
andika & verb & 11 & `write' & kam, kik, kin, kon, lin, lug, mer, nya, run, swh, xho \\
mama & noun & 11 & `mother' & kam, kik, kin, kon, lin, lug, mer, sna, swh, xho, zul \\
tuma & verb & 11 & `send' & kam, kik, kin, kon, lin, lug, mer, nso, nya, run, swh \\
linda & verb & 11 & `wait, guard' & kam, kik, kin, kon, lin, lug, mer, nso, swh, xho, zul \\
nena & verb & 11 & `speak, say' & kam, kik, kin, kon, lin, lug, mer, nya, swh, xho, zul \\
tenda & verb & 11 & `do, act' & kam, kik, kin, kon, lin, lug, mer, nya, sna, swh, zul \\
genda & verb & 11 & `go, travel' & kam, kik, kin, kon, lin, lug, mer, nya, run, swh, xho \\
teka & verb & 11 & `take, fetch' & kam, kik, kin, kon, lin, lug, mer, nso, run, sna, swh \\
muka & noun & 11 & `woman, wife' & kam, kik, kin, kon, lin, lug, mer, run, sna, swh, zul \\
paka & verb & 10 & `cat, smear' & kin, kon, lin, lug, nso, nya, run, sna, swh, xho \\
baba & noun & 10 & `father, wing' & kik, kin, kon, lin, lug, mer, run, sna, swh, zul \\
soma & verb & 10 & `read, learn' & kam, kik, kin, lug, mer, nya, run, sna, swh, zul \\
kamba & noun & 10 & `rope' & kam, kik, kin, kon, lin, lug, mer, nya, run, sna \\
\bottomrule
\end{tabular}
\end{table*}

\end{document}